\documentclass{article}
\usepackage{spconf,amsmath,graphicx}

\usepackage{amsmath}
\usepackage{tikz}
\usetikzlibrary{graphs}
\usepackage{subfigure}

\usepackage{amsfonts}

\title{Deep Transfer Learning for EEG-based Brain Computer Interface}
\name{Chuanqi Tan \qquad Fuchun Sun \qquad Wenchang Zhang\thanks{This work was supported by the China National Natural Fund: 91420302 and 91520201.
	}}
\address{State Key Laboratory of Intelligent Technology and Systems
    \\Tsinghua National Laboratory for Information Science and Technology (TNList)
    \\Department of Computer Science and Technology, Tsinghua University\\
	fcsun@tsinghua.edu.cn, \{tcq15, zhangwc14\}@mails.tsinghua.edu.cn}
\begin{document}
\maketitle
\begin{abstract}
The electroencephalography classifier is the most important component of brain-computer interface based systems. There are two major problems hindering the improvement of it. First, traditional methods do not fully exploit multimodal information. Second, large-scale annotated EEG datasets are almost impossible to acquire because biological data acquisition is challenging and quality annotation is costly.
Herein, we propose a novel deep transfer learning approach to solve these two problems. First, we model cognitive events based on EEG data by characterizing the data using EEG optical flow, which is designed to preserve multimodal EEG information in a uniform representation. Second, we design a deep transfer learning framework which is suitable for transferring knowledge by joint training, which contains a adversarial network and a special loss function.
The experiments demonstrate that our approach, when applied to EEG classification tasks, has many advantages, such as robustness and accuracy. 

\end{abstract}
\section{Introduction}

For patients suffering from stroke, it is meaningful to provide a communication method apart from the normal nerve-muscle output pathway to deliver brain messages and commands to the external world.
Due to natural and non-intrusive characteristics, most brain-computer interface (BCI) systems select electroencephalography (EEG) signals as the input \cite{amiri2013review}. The biggest challenge and most important component in a BCI system is the EEG classifier, which translates a raw EEG signal into the commands of the human brain. 

Currently, two major problems hinder the improvement of EEG classification. First, traditional EEG classification methods focus on frequency-domain information and cannot fully exploit multimodal information. Second,  high-quality, large-scale annotated EEG datasets are extremely difficult to construct because biological data acquisition is challenging and quality annotation is costly.

We solved these problems in the following ways. 
First, we modeled cognitive events based on EEG data by characterizing the data using EEG optical flow, which is designed to preserve multimodal EEG information. In this way, the EEG classification problem is reduced to a video classification problem and can be solved using advanced computer vision technology.
Second, we designed a deep transfer learning framework suitable for transferring knowledge by joint training, which contains an adversarial network and a special loss function. The architecture of the transfer network was borrowed from computer vision, and the parameters of the transfer network were jointly trained on ImageNet and EEG optical flow. In order to achieve a more efficient transfer learning, a special loss function which considered the performance of the adversarial network was used in the joint training. After that, a classification network was trained on EEG optical flow to obtain the final category label.
Although natural images and EEG signals are significantly different, EEG optical flow provides a similar representation of the two different domains, and our joint training algorithm brings the two different domains closer together.

The \textbf{contributions} of this paper are as follows: 
(1) We propose EEG optical flow, which was designed to preserve multimodal EEG information in a uniform representation, reduce the EEG classification problem to a video classification problem. 
(2) We construct a deep transfer learning framework to transfer knowledge from computer vision in a sophisticate way, which solves the problem of insufficient biological training data. 
(3) We perform experiments on public dataset, and the results show that our approach has many advantages over traditional methods.

\section{Related Work}

Substantial work has been conducted to improve EEG classification accuracy. The performance of this pattern-recognition-like system depends on both the selected features and the employed classification algorithms. 
As a new classification platform, deep learning has recently received increasing attention from researchers \cite{lecun2015deep} and has been successfully applied to many classification problems, such as image classification, video classification and speech recognition. 
Bioinformatics has also benefited from deep learning. In recent years, many public reviews \cite{Mamoshina2016Applications} have discussed deep learning applications in bioinformatics research, for example, applying deep belief networks (DBN) to the frequency components of EEG signals to classify left-hand and right-hand motor imagery skills \cite{An2014A}. \cite{Cecotti2011Convolutional} used a convolutional neural network (CNN) to decode P300 patterns. \cite{Soleymani2014Continuous} conducted an emotion detection and facial expressions study with both EEG signals and face images using a recurrent neural network (RNN).

Transfer learning enables the use of different domains, tasks, and distributions for training and testing \cite{pan2010survey}. 
\cite{jayaram2016transfer} reviewed the current state-of-the-art transfer learning approaches in BCI.
\cite{hajinoroozi2017deep} transferred general features via a convolutional network across subjects and experiments. 
\cite{zheng2016personalizing} applied kernel principle analysis and transductive parameter transfer to identify the relationships between classifier parameter vectors across subjects. 
\cite{lin2017improving} evaluated the transferability between subjects by calculating distance and transferred knowledge in comparable feature spaces to improve accuracy.
\cite{tzeng2015simultaneous} proposed an approach which can simultaneously transfer knowledge across domains and tasks. 
\cite{long2015learning} and \cite{long2016deep} attempted the learning of transferable features by embedding task-specific layers in a reproducing kernel Hilbert space where the mean embeddings of different domain distributions can be explicitly matched.

To the best of our knowledge, no researchers have attempted to transfer knowledge from computer vision to EEG classification. Large-scale, high-quality annotated datasets, such as ImageNet, and many excellent deep neural networks can greatly facilitate EEG classification.

\section{Methods}

\subsection{EEG Optical Flow}

Traditional methods \textit{do not fully exploit multimodal information}. For example, they ignore the locations of the electrodes and the inherent information in the spatial dimension. 
In our approach, we convert raw EEG signals into EEG optical flow to represent the multimodal information of EEG.

\begin{figure}[h]
    \centering
    \subfigure{
        \includegraphics[width=0.65in]{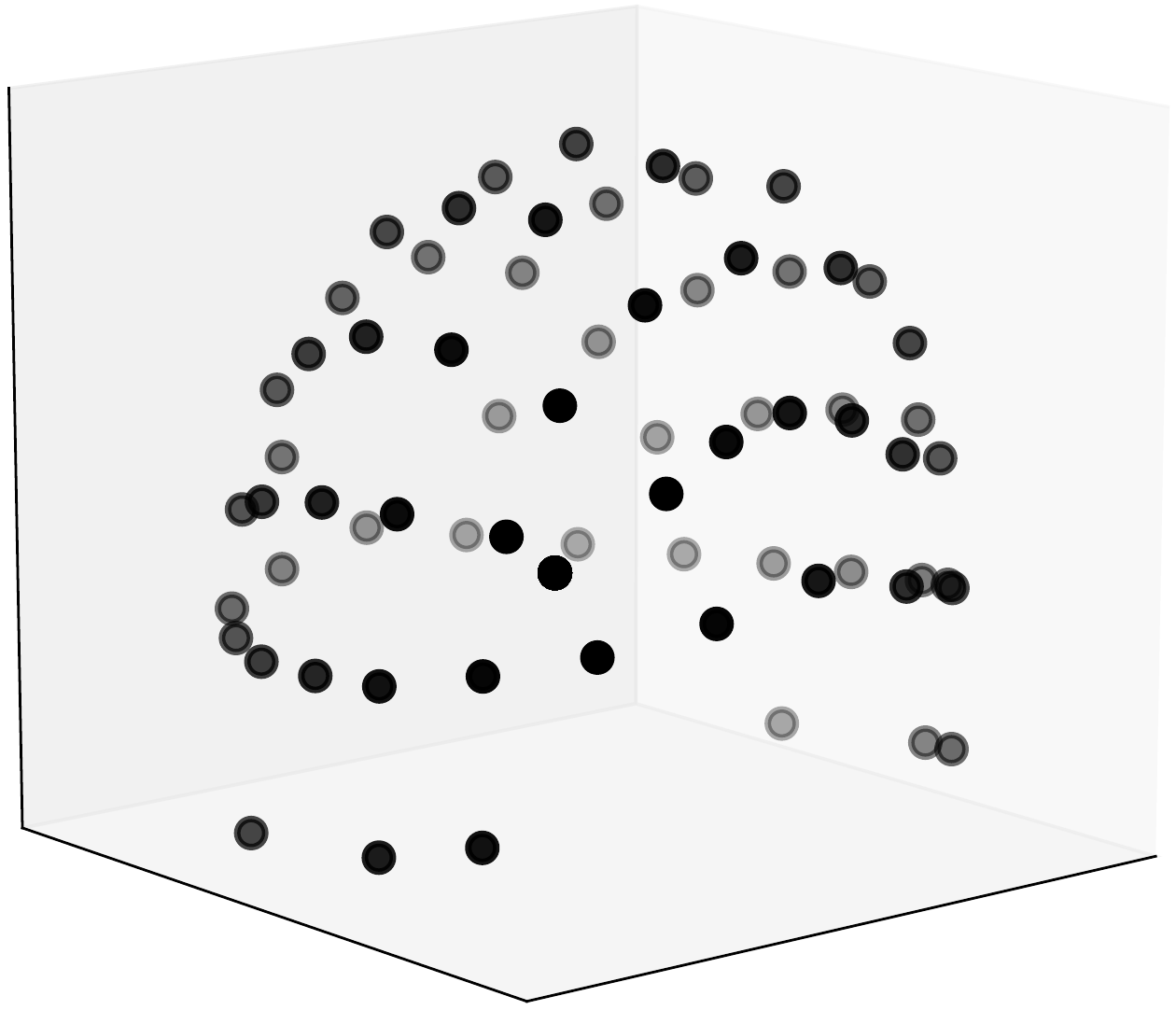}
        \label{project to eeg image:a}
    }
    \hspace{0.2in}
    \subfigure{
        \includegraphics[width=0.56in]{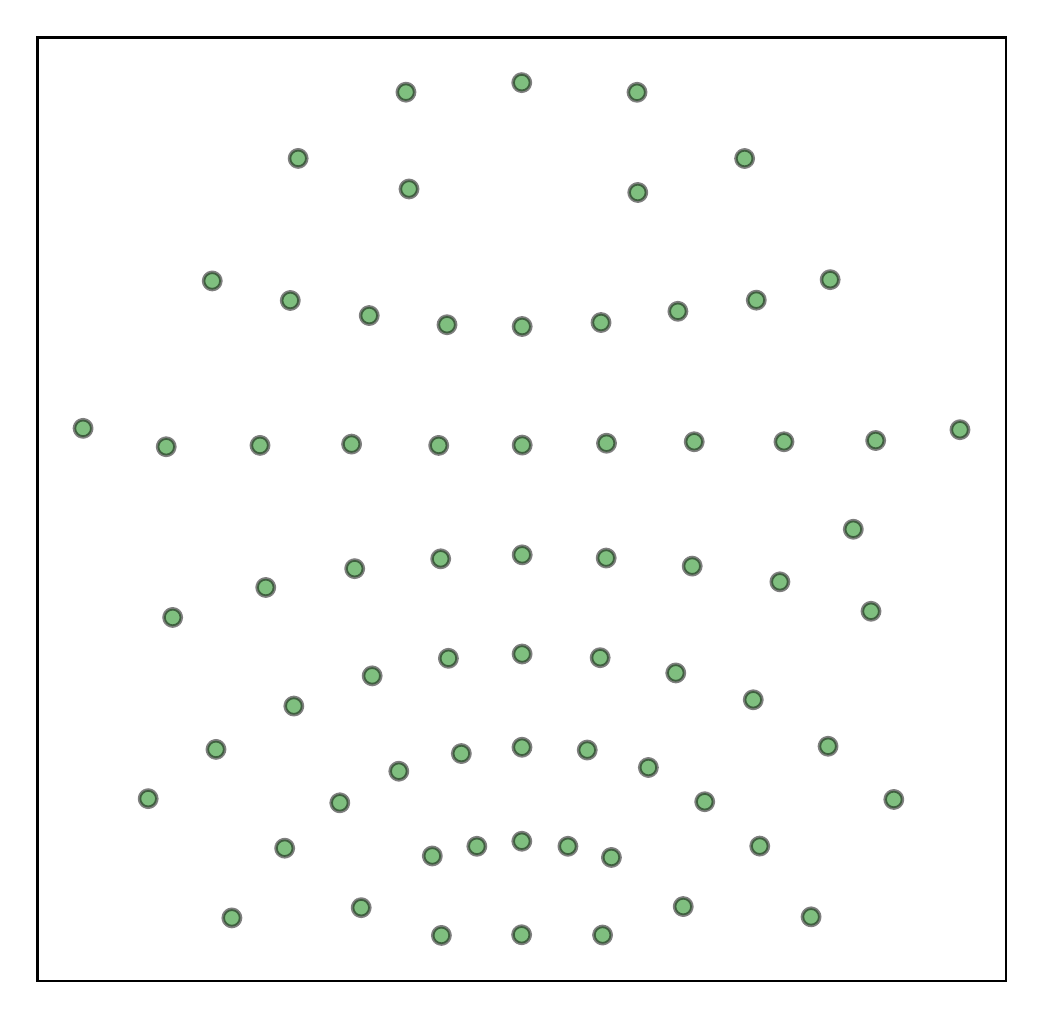}
        \label{project to eeg image:b}
    }
    \hspace{0.2in}
    \subfigure{
        \includegraphics[width=0.52in]{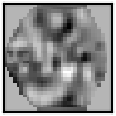}
        \label{project to eeg image:c}
    }
    \caption{Generating EEG video by projecting and interpolating.}
    \label{project to eeg image}
\end{figure}

\textbf{EEG video} is converted from the raw EEG signal. First, filtering is performed using five stereotyped frequency filters ($\alpha$: 8-13 Hz, $\beta$: 14-30 Hz, $\gamma$: 31-51 Hz, $\delta$: 0.5-3 Hz, $\theta$: 4-7 Hz) to characterize different EEG signal rhythms. 
Second, EEG video frames are generated from each raw EEG signal frame in the time dimension. We project the 3D locations of the electrodes to 2D points via azimuthal equidistant projection (AEP), which borrows from mapping applications, and interpolate them to gray image by clough-tocher algorithm. The processes are shown in Figure \ref{project to eeg image}.
AEP can maintain the distance between electrodes more proportionately to represent more useful information in the spatial dimension.

\textbf{EEG optical flow} is extracted from the converted EEG video. Optical flow \cite{Farneb2003Two} is introduced in our approach to describe the variant information of the EEG signal. Optical flow is widely used in video classification approaches because it can describe the obvious motion of objects in a visual scene by calculating the motion between two neighboring image frames taken at times $t$ and $t+\Delta t$ at every pixel position.

\begin{figure}[h]
    \centering
    \begin{tikzpicture}
    \node[anchor=south west,inner sep=0] (a) at (0.1,0.1) {\includegraphics[width=0.6in]{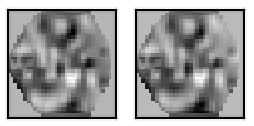}};
    \draw [blue, dashed, very thick] (0,0) rectangle (1.7,0.9);
    \node[anchor=south west,inner sep=0] (c) at (3,0.1) {\includegraphics[width=0.3in]{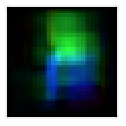}};
    
    \draw[->,>=stealth,blue,line width=1mm] (a) to (c);
    \end{tikzpicture}
    
    \caption{Visualization of the EEG optical flow frame.}
    \label{generate optical flow}
\end{figure}

We store the optical flow as an image and rescale it to [0, 255]. The visualization image is shown in Figure \ref{generate optical flow} as the mapping direction and magnitude to an HSV image.

\begin{figure*}[t]
    \centering
    \setlength{\fboxsep}{0cm} 
    \fbox{\includegraphics[width=6.5in]{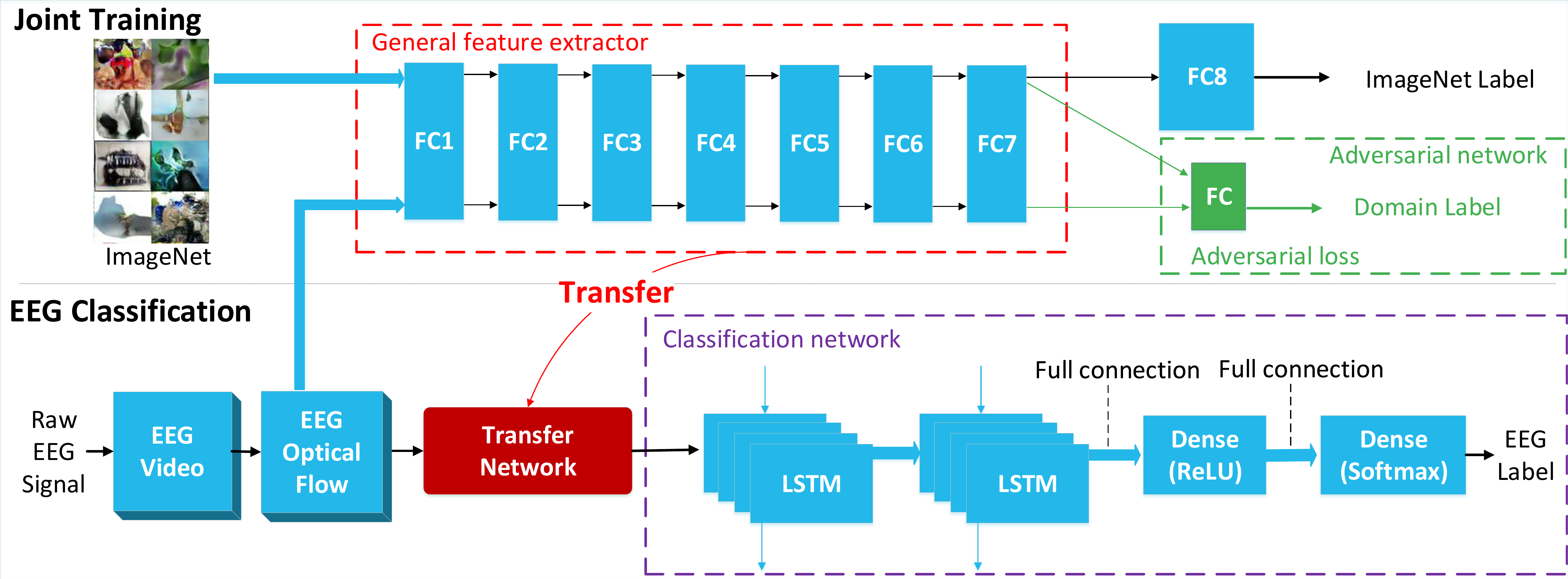}}
    \caption{Architecture of our deep transfer learning framework, using AlexNet as an example of transfer network.
    }
    \label{network architecture}
\end{figure*}

Many \textbf{benefits} can be gained from using the EEG optical flow. 
\emph{Uniform representation of multimodal information}: The spatial structure of the electrodes is preserved by the AEP, and the spectral information extracted via five stereotyped frequency filters and the temporal information are represented by the optical flow.
\emph{Suitable for CNN}: Due to the inherent structure of CNNs, the EEG optical flow is more compatible to the image and video data structure. CNN can discover the regional information of EEG optical flow, which reflects the regional information of brain regions.
\emph{Transfer learning ability}: By reducing the EEG classification problem to a video classification problem, we gain the ability to transfer knowledge from computer vision, which has large-scale annotated datasets, such as ImageNet, and many excellent networks.

The entire dataset can be divided into independent epochs, which are the responses to stimulus events, according to the label of the stimulus channel. 
We employ a resampling algorithm to extend the training data, which plays a positive role in improving the generalization performance of the classifier, which we discuss in the following section.

\subsection{Network Architecture and Transfer Learning}

\emph{Insufficient training data} is a serious problem in all domains related to bioinformatics. 
Our answer to this problem is transfer learning, which transfers knowledge from computer vision. 
We construct a deep transfer learning framework that contains two steps to obtain the final EEG category labels. 
The architecture of our network is shown in Figure \ref{network architecture}.

\textbf{Joint training} aimed to learn a better representation for natural images and EEG optical flow. Many studies have demonstrated that front layers in an CNN network can extract the \textit{general features} of images, such as edges and corners (shown in the red box). But the general feature extractor trained by natural images does not fully match the EEG optical flow.
Many previous works have shown that in order to achieve a better effect of transfer, the edge distribution of features from the source domain and target domain should be as similar as possible. 
Inspired by generative adversarial nets (GAN), we apply an adversarial network (shown in the green box) to train a better general feature extractor. We use features extracted from natural images and EEG optical flow as the inputs for the adversarial network and train it to identify their origins. If the adversarial network achieves worse performance, it means a small difference between the two types of feature and  better transferability, and vice versa.

Suppose we have ImageNet $\{X_{img}, Y_{img}\}$ and EEG optical flow $\{X_{of}, Y_{of}\}$, where $Y$ is labels of data. Our goal of joint training is to produce a general feature extractor with parameters $\theta_{extr}$ that can represent natural images and EEG optical flow in a suitable way and can correctly classify in the future. $\theta_{img}$ and $\theta_{adver}$ points to parameters of the ImageNet classification network and the adversarial network.

In order to extract more transferable general features, we use a special loss function while ImageNet classification training, which takes into account the performance of the adversarial network.
It can be defined as follows:

\begin{eqnarray}
\label{eq:loss1} && \mathcal{L} _{img} = -\sum_{k}{ \mathbb{I} [y=k]\log p_{k}} + \alpha \mathcal L_{adver} \\
\label{eq:loss2} && \mathcal L_{adver} = -\sum_{d}{\frac{1}{D} \log p_{d}} 
\end{eqnarray}

where $k$ is the number of categories, $p_{k}$ is the softmax of the classifier activations, 
$\mathcal L_{adver}$ is the cross entropy of the adversarial network and $\alpha$ is the hyper parameter of how strongly $\mathcal L_{adver}$ influences the optimization.

This loss function will minimize the performance of the adversarial network. It is means that the two types of feature will have similar edge distributions as possible, that is not easily distinguish by the adversarial network.

While training the ImageNet classification network, the loss function wants to reduce the performance of the adversarial network by optimizing Equation (\ref{eq:iterator1}).
But while training the adversarial network, it tries to improve the performance of the adversarial network by optimizing Equation (\ref{eq:iterator2}). 
These two goals stand in direct opposition to one another, and we overcome this by iteratively optimizing the following two goals while fixing other parameters.

\begin{eqnarray}
\label{eq:iterator1} & \underset{\theta_{extr} \theta_{img}} {\mathrm{argmin}} {\mathcal{L} _{img} (X_{img}, X_{of}, \theta_{adver}; \theta_{extr}, \theta_{img})}  \\
\label{eq:iterator2} & \underset{\theta_{extr} \theta_{adver}} {\mathrm{argmin}} {\mathcal{L} _{adver} (X_{img}, X_{of}, \theta_{img}; \theta_{extr}, \theta_{adver})} 
\end{eqnarray}

Joint training force the transfer network to discover general features with more transferability, which is important to obtain useful knowledge from natural images and transfer it to EEG optical flow.

\textbf{EEG classification} aimed to obtain the final EEG label. General features are extracted by transfer network and pre-trained parameters (identified by the red arrow).
Then features are fed to a classification network (shown in the purple box) with two RNN layers and two fully connected layers. We use long-short term memory (LSTM) to prevent vanishing gradient problems in the time dimension when training. Two fully connected layers are applied at the end of classification network, with the last layer applying a softmax activation function to obtain the final EEG label.
If a fine-tuning strategy is used, the last one or two layers of the transfer network will be updated simultaneously in the EEG Classification.

\section{Experiments}

We apply our approach to a famous public dataset in the BCI field, named Open Music Imagery Information Retrieval (OpenMIIR) published by the Brain and Mind Institute at the University of Western Ontario \cite{stober2015towards,stober2017learning}.
The parameters used in our approach are described as follows. We convert raw EEG signals into EEG videos with thirteen frames and a 32*32 resolution. These frames are resampled 50 times and converted to EEG optical flow with twelve frames.
In the classification network, the recurrent layers contain 128 nodes. After the LSTM layers are applied, a dropout layer with a 0.25 ratio is applied to disable a portion of the neurons. The fully connected layers in the EEG classification network contains 64 nodes.
We employ many popular, excellent networks as the target of transfer learning. These networks, including AlexNet, VGG16, VGG19 and ResNet, are the winners of past ImageNet Large Scale Visual Recognition Competition (ILSVRC) competitions \cite{guo2016deep}.

\subsection{Results}

\begin{figure}[h]
    \centering 
    \begin{minipage}[h]{\linewidth} 
        \centering 
        
        \subfigure[AlexNet]{
            \includegraphics[width=0.72in]{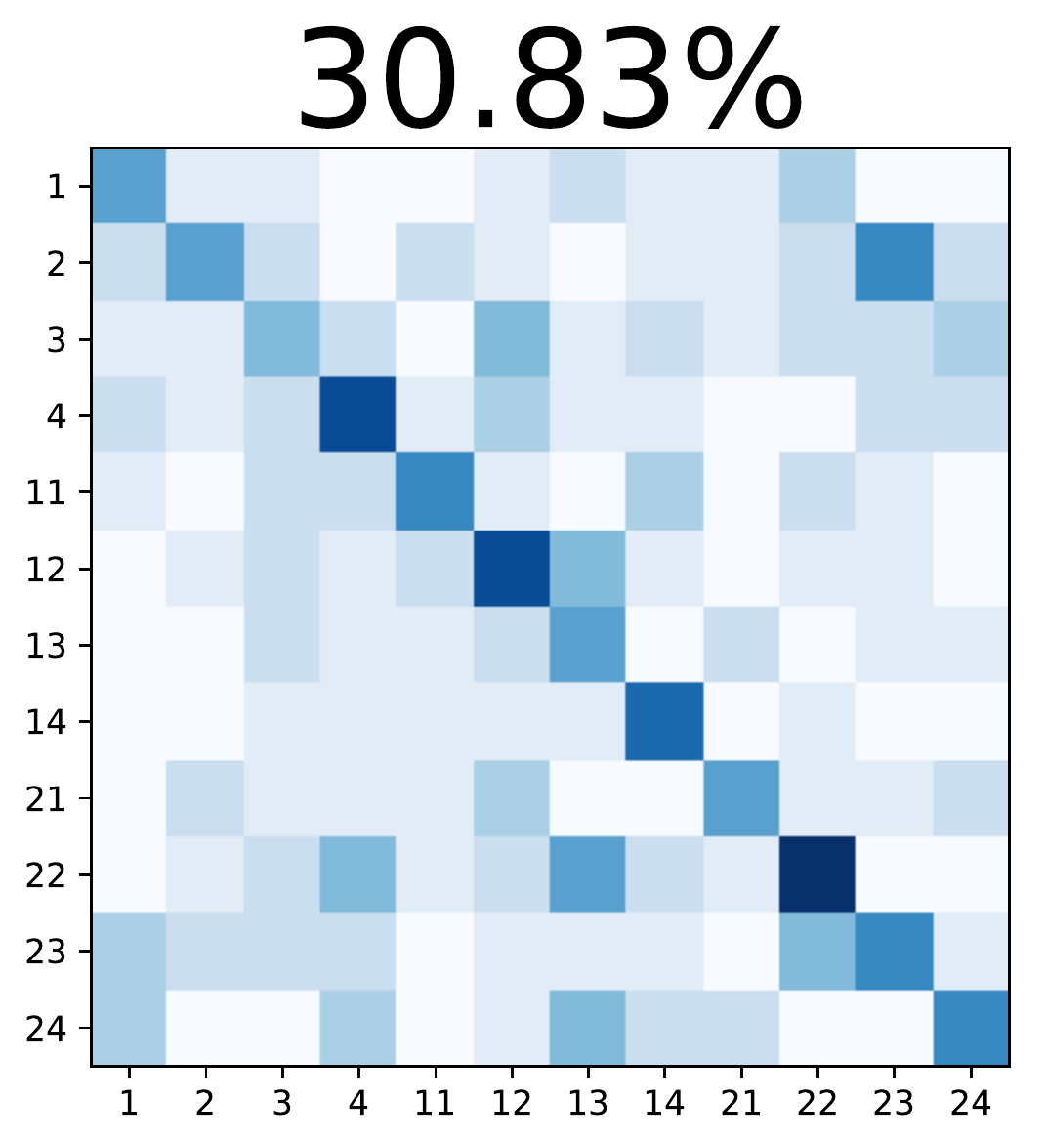}
        }
        \hspace{0in}
        \subfigure[VGG16]{
            \includegraphics[width=0.72in]{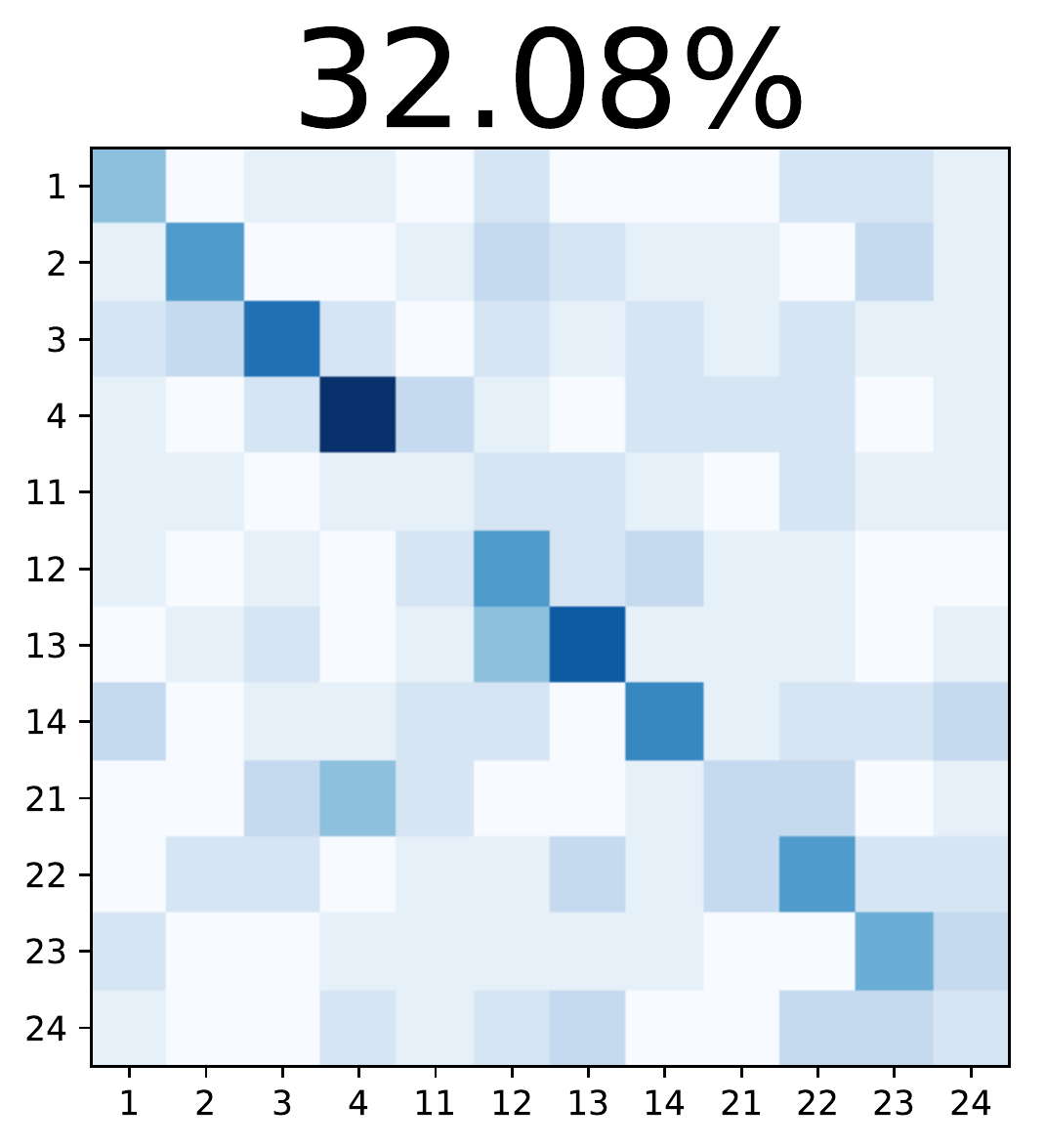}
        }
        \hspace{0in}
        \subfigure[VGG19]{
            \includegraphics[width=0.72in]{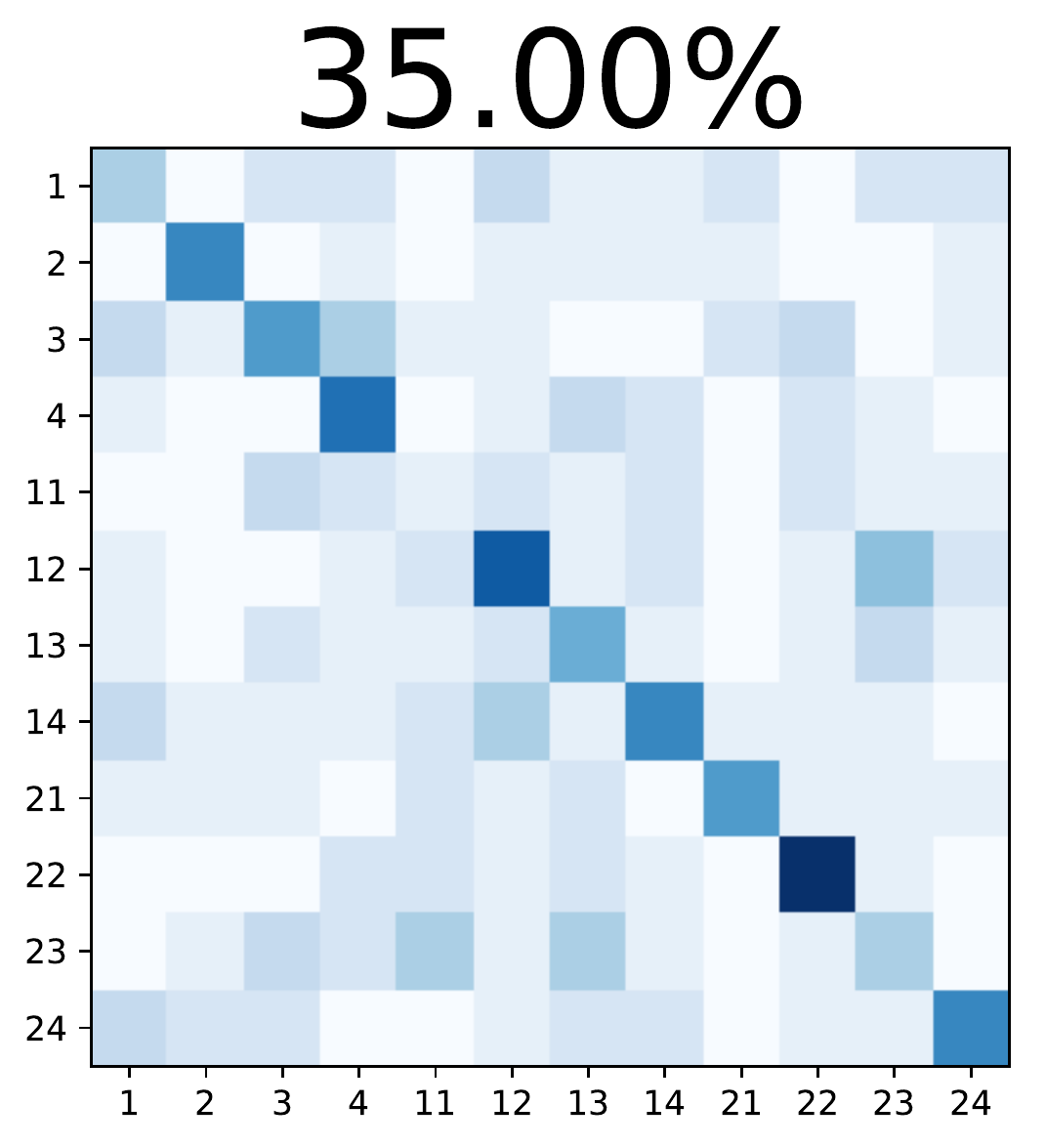}
        }
        \hspace{0in}
        \subfigure[ResNet]{
            \includegraphics[width=0.72in]{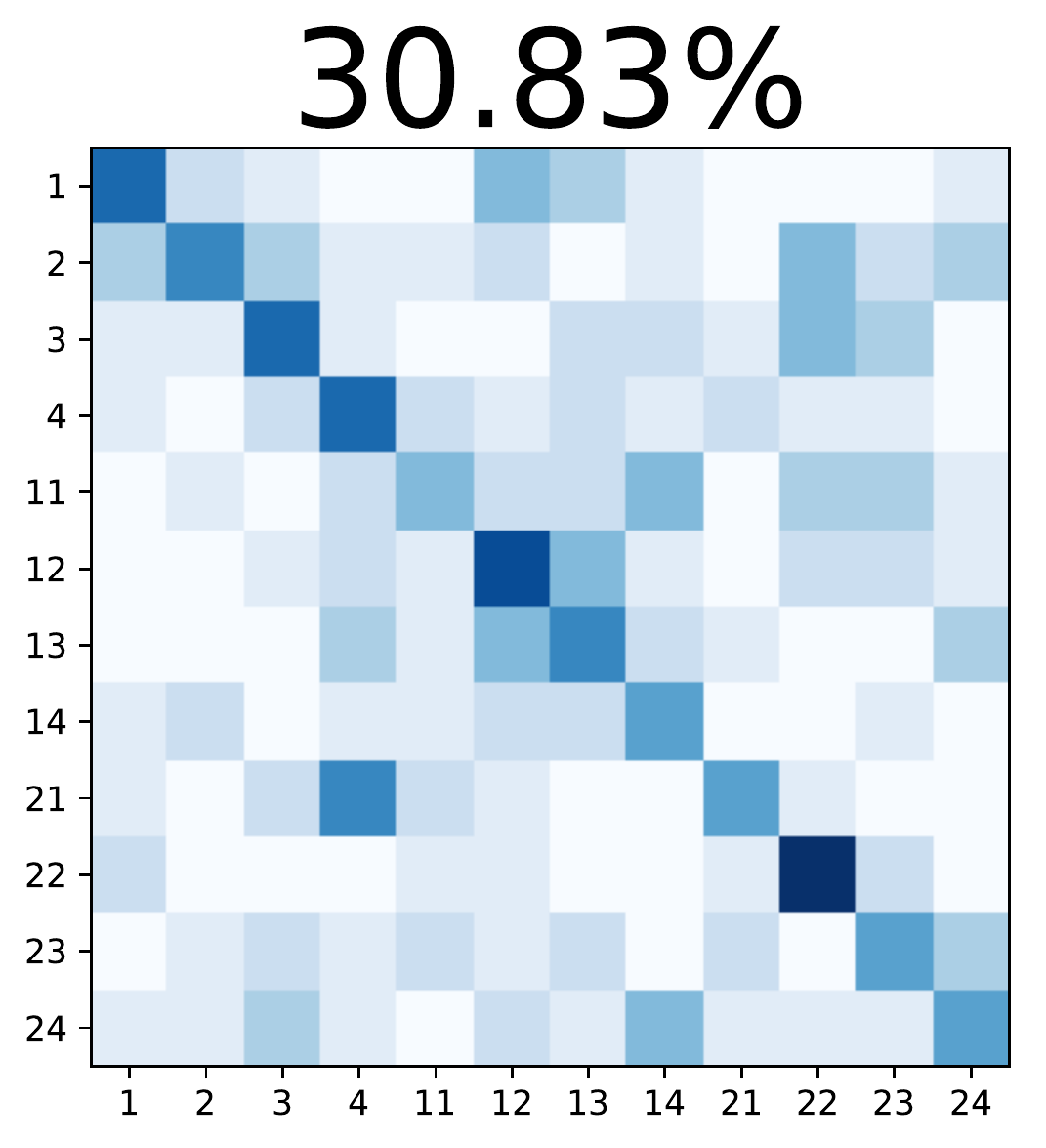}
        }
    \end{minipage}%
    
    \caption{Confusion matrix of our approach with different transfer networks used in experiments.}
    \label{figure:confusion matrix of OpenMIIR}
\end{figure}

According to the approach described in the previous sections, we carry out classification experiments on the dataset OpenMIIR. 
The OpenMIIR dataset does not distinguish between training and test sets, therefore, we randomly selected 10\% of the dataset to be used as the test dataset.
The experimental results show that our approach is superior to other current state-of-the-art methods with better classification accuracy. Figure \ref{figure:confusion matrix of OpenMIIR} shows the 12-class confusion matrix of the experimental results when different transfer networks are used for our deep transfer learning framework.

One important goal of our approach is solve the problem of insufficient training data. 
Due to the design of our approach, it is possible to train a large-scale deep neural network using a limited EEG training dataset by transferring knowledge from computer vision. 
To determine the utility of our solution, we test our approach while further reducing the training dataset. 
As the baseline, we tested three recent proposed methods: the support vector machine classifier (SVC) described in \cite{fan2008liblinear}, the deep neural network (DNN) described in \cite{stober2017learning} and the CNN described in \cite{stober2015deep}.
Experiments on the dataset OpenMIIR are carried out to compare the performance of our approach and the baseline methods while further reducing the training dataset, and the results are shown in Table \ref{table: reduce_training_data}.
In addition, we tested our approach without joint training.

\begin{table}[h]
    \centering
    \caption{Classification accuracy while further reducing the size of the training dataset. $N\%$ in the table head indicate the percentage of training dataset were used during training. The values in parentheses indicate the classification results without joint training.}
    \label{table: reduce_training_data}
    \begin{tabular}{llllllllllll}
        \hline\noalign{\smallskip}
        ~ & $100\%$ & $50\%$ & $25\%$ \\ \noalign{\smallskip}
        \hline
        \noalign{\smallskip}
        SVC & 23.1 & 16.69 & 9.83 \\
        DNN & 27.22 & 20.83 & 12.47 \\
        CNN & 27.8 & 19.2 & 8.55 \\
        ${AlexNet}$ & 30.83(27.92) & 24.58(21.72) & \textbf{15.42}(15.27) \\
        ${VGG16}$ & 32.08(31.67) & \textbf{28.33}(26.2) & \textbf{15.42}(14.7) \\
        ${VGG19}$ & \textbf{35.0}(32.92) & 25.83(23.14) & 14.58(13.93) \\
        ${ResNet}$ & 30.83(27.08) & 22.5(22.67) & 11.25(10.42) \\
        \hline
    \end{tabular}
\end{table}

\subsection{Discussion}

We can draw the following conclusions from the experimental results presented in the previous section:
(1) The experimental results shown in Figure \ref{figure:confusion matrix of OpenMIIR} and Table \ref{table: reduce_training_data} demonstrate that our proposed approach achieves accuracy that is obviously superior to that of the traditional methods;
(2) VGG16 and VGG19 are good choices of transfer network;
(3) Table \ref{table: reduce_training_data} shows that our approach can achieve acceptable results while further reducing the size of the training set;
(4) Joint training play a important and positive role in the final results. 
Due to the benefits of transfer learning, we can train these large neural networks with a limited training dataset. However, compared to the traditional EEG classification approaches, our network requires more time for prediction due to the network complexity, which may present obstacles when our approach is applied to real-time BCI systems.

\section{Conclusions}

In this paper, we propose a novel EEG signal classification approach with EEG optical flow and deep transfer learning in response to two major problems in EEG classification: (1) the inability of traditional methods to fully exploit multimodal information and (2) insufficient training data.
Our approach is superior to other state-of-the-art methods, which is important for building better BCI systems, and provides a new perspective for solving the problem of EEG classification.
In the future, we plan to develop an improved network based on state-of-the-art methods in computer vision and other domains. 
In addition, our approach can be viewed as a general bioelectrical signal classification framework that is suitable for other bioelectrical signals, such as functional magnetic resonance imaging (fMRI).

\bibliographystyle{IEEEbib}
\bibliography{icassp_reference}

\begin{thebibliography}{10}

\bibitem{amiri2013review}
Setare Amiri, Reza Fazel-Rezai, and Vahid Asadpour,
\newblock ``A review of hybrid brain-computer interface systems,''
\newblock {\em Advances in Human-Computer Interaction}, vol. 2013, pp. 1, 2013.

\bibitem{lecun2015deep}
Yann LeCun, Yoshua Bengio, and Geoffrey Hinton,
\newblock ``Deep learning,''
\newblock {\em Nature}, vol. 521, no. 7553, pp. 436--444, 2015.

\bibitem{Mamoshina2016Applications}
P~Mamoshina, A~Vieira, E~Putin, and A~Zhavoronkov,
\newblock ``Applications of deep learning in biomedicine.,''
\newblock {\em Molecular Pharmaceutics}, vol. 13, no. 5, pp. 1445, 2016.

\bibitem{An2014A}
Xiu An, Deping Kuang, Xiaojiao Guo, Yilu Zhao, and Lianghua He,
\newblock ``A deep learning method for classification of eeg data based on
  motor imagery,''
\newblock in {\em International Conference on Intelligent Computing}, 2014, pp.
  203--210.

\bibitem{Cecotti2011Convolutional}
Hubert Cecotti and Axel Graser,
\newblock ``Convolutional neural networks for p300 detection with application
  to brain-computer interfaces,''
\newblock {\em IEEE Transactions on Pattern Analysis \& Machine Intelligence},
  vol. 33, no. 3, pp. 433, 2011.

\bibitem{Soleymani2014Continuous}
Mohammad Soleymani, Sadjad Asghariesfeden, Maja Pantic, and Yun Fu,
\newblock ``Continuous emotion detection using eeg signals and facial
  expressions,''
\newblock in {\em IEEE International Conference on Multimedia and Expo}, 2014,
  pp. 1--6.

\bibitem{pan2010survey}
Sinno~Jialin Pan and Qiang Yang,
\newblock ``A survey on transfer learning,''
\newblock {\em IEEE Transactions on knowledge and data engineering}, vol. 22,
  no. 10, pp. 1345--1359, 2010.

\bibitem{jayaram2016transfer}
Vinay Jayaram, Morteza Alamgir, Yasemin Altun, Bernhard Scholkopf, and Moritz
  Grosse-Wentrup,
\newblock ``Transfer learning in brain-computer interfaces,''
\newblock {\em IEEE Computational Intelligence Magazine}, vol. 11, no. 1, pp.
  20--31, 2016.

\bibitem{hajinoroozi2017deep}
Mehdi Hajinoroozi, Zijing Mao, Yuan-Pin Lin, and Yufei Huang,
\newblock ``Deep transfer learning for cross-subject and cross-experiment
  prediction of image rapid serial visual presentation events from eeg data,''
\newblock in {\em International Conference on Augmented Cognition}. Springer,
  2017, pp. 45--55.

\bibitem{zheng2016personalizing}
Wei-Long Zheng and Bao-Liang Lu,
\newblock ``Personalizing eeg-based affective models with transfer learning,''
\newblock in {\em Proceedings of the Twenty-Fifth International Joint
  Conference on Artificial Intelligence}. AAAI Press, 2016, pp. 2732--2738.

\bibitem{lin2017improving}
Yuan-Pin Lin and Tzyy-Ping Jung,
\newblock ``Improving eeg-based emotion classification using conditional
  transfer learning,''
\newblock {\em Frontiers in Human Neuroscience}, vol. 11, 2017.

\bibitem{tzeng2015simultaneous}
Eric Tzeng, Judy Hoffman, Trevor Darrell, and Kate Saenko,
\newblock ``Simultaneous deep transfer across domains and tasks,''
\newblock in {\em Proceedings of the IEEE International Conference on Computer
  Vision}, 2015, pp. 4068--4076.

\bibitem{long2015learning}
Mingsheng Long, Yue Cao, Jianmin Wang, and Michael Jordan,
\newblock ``Learning transferable features with deep adaptation networks,''
\newblock in {\em International Conference on Machine Learning}, 2015, pp.
  97--105.

\bibitem{long2016deep}
Mingsheng Long, Jianmin Wang, and Michael~I Jordan,
\newblock ``Deep transfer learning with joint adaptation networks,''
\newblock {\em arXiv preprint arXiv:1605.06636}, 2016.

\bibitem{Farneb2003Two}
Farneb and Gunnar Ck,
\newblock ``Two-frame motion estimation based on polynomial expansion,''
\newblock in {\em Scandinavian Conference on Image Analysis}, 2003, pp.
  363--370.

\bibitem{stober2015towards}
Sebastian Stober, Avital Sternin, Adrian~M Owen, and Jessica~A Grahn,
\newblock ``Towards music imagery information retrieval: Introducing the
  openmiir dataset of eeg recordings from music perception and imagination.,''
\newblock in {\em ISMIR}, 2015, pp. 763--769.

\bibitem{stober2017learning}
Sebastian Stober,
\newblock ``Learning discriminative features from electroencephalography
  recordings by encoding similarity constraints,''
\newblock in {\em Acoustics, Speech and Signal Processing (ICASSP), 2017 IEEE
  International Conference on}. IEEE, 2017, pp. 6175--6179.

\bibitem{guo2016deep}
Yanming Guo, Yu~Liu, Ard Oerlemans, Songyang Lao, Song Wu, and Michael~S Lew,
\newblock ``Deep learning for visual understanding: A review,''
\newblock {\em Neurocomputing}, vol. 187, pp. 27--48, 2016.

\bibitem{fan2008liblinear}
Rong-En Fan, Kai-Wei Chang, Cho-Jui Hsieh, Xiang-Rui Wang, and Chih-Jen Lin,
\newblock ``Liblinear: A library for large linear classification,''
\newblock {\em Journal of machine learning research}, vol. 9, no. Aug, pp.
  1871--1874, 2008.

\bibitem{stober2015deep}
Sebastian Stober, Avital Sternin, Adrian~M Owen, and Jessica~A Grahn,
\newblock ``Deep feature learning for eeg recordings,''
\newblock {\em arXiv preprint arXiv:1511.04306}, 2015.

\end{thebibliography}

\end{document}